\documentclass[letterpaper,twocolumn,10pt]{article}
\usepackage{usenix2019_v3}

\usepackage{tikz}
\usepackage{amsmath}
\usepackage{amssymb}
\usepackage{booktabs}
\usepackage{filecontents}
\usepackage{graphicx}
\usepackage{subcaption}
\usepackage[most]{tcolorbox}
\tcbuselibrary{breakable}

\usepackage{xcolor}
\definecolor{linkteal}{RGB}{159,29,42}
\usepackage{hyperref}
\usepackage{fontawesome5}
\usepackage{graphicx}

\hypersetup{
    colorlinks=true,
    urlcolor=black
}
\usepackage{eso-pic}
\newtcolorbox{takeawaybox}{
    enhanced,
    breakable,
    colback=gray!6,
    colframe=gray!25,
    boxrule=0.5pt,
    arc=2pt,
    left=5pt,
    right=5pt,
    top=4pt,
    bottom=4pt,
    fonttitle=\bfseries,
    title=Takeaway
}

\begin{document}

\date{}

\title{\Large \bf BadWAM: When World-Action Models Dream Right but Act Wrong}

\author{
{\rm Qi Li}$^{1}$ \quad
{\rm Xingyi Yang}$^{2}$ \quad
{\rm Xinchao Wang}$^{1\dagger}$ \\
$^{1}$National University of Singapore
\quad
$^{2}$The Hong Kong Polytechnic University \\
{\tt liqi@u.nus.edu}
\quad
{\tt xinchao@nus.edu.sg} \\
{\small $^\dagger$Corresponding author.}\\[0.8ex]
{\small
\href{https://github.com/LiQiiiii/BadWAM}
{\faGithub\ GitHub}
\quad\quad
\href{https://liqiiiii.github.io/BadWAM}
{\faGlobe\ Homepage}
\quad\quad
\href{https://huggingface.co/collections/LIQIIIII/badwam}
{\faRobot\ Hugging Face}
} 
}

\maketitle

\begin{abstract}
World-action models (WAMs) are emerging as a promising foundation for embodied control: rather than predicting actions alone, they learn representations that couple action generation with future world prediction. This coupling is often viewed as a source of robustness, interpretability, and safety, as a robot's action can in principle be checked against its imagined future. In this paper, we show that this assumption is fragile. We introduce \emph{BadWAM}, a unified framework for modeling and evaluating \emph{World-Action Drift Attacks}: a new class of WAM-specific adversarial attacks that use small visual perturbations to break the alignment between what a WAM imagines and what it executes. BadWAM characterizes this attack surface along two natural criteria: attack strength and stealthiness. When the adversary prioritizes disruption, BadWAM instantiates an \emph{action-only adversarial attack}, which directly drives the model toward task-failing actions. When the adversary additionally prioritizes stealth, BadWAM instantiates an \emph{imagination-preserving adversarial attack}, which seeks to induce harmful action shifts while keeping the model's predicted future close to its clean imagination. Together, these two attacks capture a spectrum of WAM-specific failures: from overt action hijacking to stealthier cases where the model appears to imagine a plausible future but executes a desynchronized action. We evaluate BadWAM across different variants of WAMs. Results show that our attacks substantially reduce task success rates under closed-loop execution. For example, our action-only attack reduces the model performance from 96.5\% to 43.1\% success. The results of our imagination-preserving attack further exposes a WAM-specific vulnerability: moderate future-preserving regularization can maintain strong attack performance while reducing future imagination drift.
\end{abstract}

\section{Introduction}

Robotic foundation models are rapidly moving from action prediction toward \emph{world-action modeling}~\cite{kim2025openvla,black2024pi_0,pi0.7,zitkovich2023rt,yuan2026fastwam,shen2026world}. A world-action model (WAM) does not merely map observations and language instructions to low-level controls; it also learns representations that couple action generation with future world prediction~\cite{yuan2026fastwam,shen2026world}. This design is attractive for embodied AI. Future prediction can provide richer physical representations, enable planning or verification, and offer an intuitive safety signal: if a robot can imagine the consequences of its actions, then unsafe or inconsistent behavior might be detected before execution~\cite{shen2026world,yuan2026fastwam,li2026vision}.

Recent WAMs instantiate this idea in different ways. Some models explicitly generate future observations or latent futures before action prediction, while others use video modeling as an auxiliary training objective and skip explicit future generation at test time~\cite{yuan2026fastwam}. Newer WAM variants further extend this paradigm toward object-addressable manipulation, mobile manipulation, and visual-tactile contact-rich control~\cite{liu2026oawam,chen2026abot,tian2026vtwam}. Across these designs, the central promise is the same: coupling action with world prediction should make robot behavior more physically grounded. This promise also motivates emerging WAM safety mechanisms that inspect imagined futures, latent rollouts, or visual trajectories to detect unsafe plans~\cite{liu2026jailwam,chen2026attackingtrusted}.

This paper begins with a basic security question: \emph{are WAMs robust to adversarially induced execution failures?} We find
that WAM coupling does not guarantee robustness: bounded visual perturbations can reduce task success across WAM variants. More specifically, \emph{action generation and future prediction can fail asynchronously}: the action output can shift substantially even when the predicted future remains close to its clean counterpart. This failure mode is qualitatively different from standard adversarial attacks on image classifiers, visual world models, or reactive robot policies~\cite{szegedy2014intriguing,goodfellow2015explaining,madry2018towards,shen2026badworld}. The attacker is not merely changing a label or shifting an action vector; it is exploiting the interface between two coupled capabilities: imagining and acting.

We call this vulnerability \emph{World-Action Drift Attack}. Rather than treating action and imagination as independent outputs, BadWAM focuses on their alignment. A bounded visual perturbation can shift the model toward task-failing behavior while leaving the predicted future comparatively close to its clean prediction. This creates a dangerous gap: a robot may appear to imagine a reasonable future, yet execute an action that is no longer aligned with that imagination. In such cases, future prediction alone is not a reliable safety signal.

We instantiate this idea in \emph{BadWAM}, a unified framework for modeling and evaluating world-action drift attacks under black-box access to a deployed WAM. BadWAM characterizes the attack surface along two natural criteria: attack strength and stealthiness. When the adversary prioritizes disruption, BadWAM instantiates an \emph{action-only adversarial attack}, which drives the model toward task-failing actions using only queryable model outputs. This attack captures the overt end of the threat spectrum, where the adversary aims to maximize execution failure.
When the adversary additionally prioritizes stealth, BadWAM instantiates an \emph{imagination-preserving adversarial attack}. This attack also operates through black-box queries, but optimizes for harmful action shifts while keeping the model's predicted future close to its clean imagination. Intuitively, this creates a stealthier failure: the robot appears to imagine a plausible future, but the action it executes has been desynchronized from that future. Together, they expose a spectrum of black-box WAM failures, from overt action hijacking to imagination-preserving failure.

We evaluate BadWAM on LIBERO~\cite{liu2023libero} and RoboTwin~\cite{chen2025robotwin}. Our experiments show that WAMs are vulnerable across variants. The action-only attack substantially reduces closed-loop task success, and the imagination-preserving attack exposes a distinct WAM-specific risk: moderate future-preserving regularization can maintain strong attack performance while reducing imagination drift. In addition, BadWAM transfers effectively across WAM variants, performing within 2.0--4.7 percentage points of attacks optimized directly on the target. This is security-critical: the vulnerability generalizes beyond a specific action head, allowing an attacker to compromise related WAM variants without target-specific optimization.

This work makes the following contributions:

\begin{itemize}
    \item We identify \emph{World-Action Drift Attack} as a WAM-specific attack surface, where action outputs can be adversarially hijacked while the model's predicted futures remain comparatively close to their clean counterparts.
    \item We design \emph{BadWAM}, a unified framework with two attack instantiations: \emph{action-only adversarial attacks} for maximizing execution disruption, and \emph{imagination-preserving adversarial attacks} for causing failures while preserving plausible predicted futures.
    \item We evaluate BadWAM on closed-loop robot manipulation tasks, showing substantial task degradation under bounded visual perturbations. We also provide detailed analyses of attack dynamics, action-channel shifts, horizon-level effects, imagination preservation, and future-weight trade-offs, offering insight into why WAMs fail under world-action drift attacks.
\end{itemize}

\section{Related Work}

\subsection{World Action Models}

World-action models have recently emerged as a predictive-action paradigm for embodied AI~\cite{shen2026world}. Unlike conventional visuomotor policies that directly map observations and instructions to actions, WAMs expose future prediction as part of the action-generation process. Let $o_t$ denote the current visual observation, $g$ denote a language instruction or goal, and $a_{t:t+H-1}$ denote an action chunk over horizon $H$. A conventional reactive policy can be abstracted as
\begin{equation}
    a_{t:t+H-1} \sim \pi_\theta(o_t, g).
\end{equation}
By contrast, a WAM additionally models future world states, such as latent future states $z_{t+1:t+K}$ or decoded future frames $v_{t+1:t+K}$. Abstractly, its world-prediction module $W_\theta$ first imagines a future, and its action module $A_\theta$ then produces an action sequence conditioned on that imagination:
\begin{equation}
    z_{t+1:t+K} \sim W_\theta(o_t, g),
    \quad
    a_{t:t+H-1} \sim A_\theta(o_t, g, z_{t+1:t+K}).
\end{equation}
In fully joint formulations, the model may instead learn a joint distribution over future states and actions:
\begin{equation}
    (z_{t+1:t+K}, a_{t:t+H-1})
    \sim
    p_\theta(z_{t+1:t+K}, a_{t:t+H-1} \mid o_t, g).
\end{equation}
In all cases, the defining feature is that action prediction is coupled, explicitly or implicitly, with a learned model of how the world may evolve.

This coupling can appear in multiple architectural forms. Some WAMs first imagine a future and then infer actions from that imagined future. Others jointly predict actions and future representations in a shared sequence model. Still others use future prediction primarily as a training signal, so that the model benefits from world modeling even if explicit future generation is not used at inference time~\cite{yuan2026fastwam}. Fast-WAM studies this design space and shows that video modeling during training can provide substantial benefits even when test-time future generation is removed~\cite{yuan2026fastwam}.
This observation is important for security: even when a model does not expose future videos at deployment time, the learned action representation may still inherit vulnerabilities from the underlying world-action coupling.

Recent WAM variants further broaden the design space. OA-WAM introduces object-addressable world representations for robust manipulation~\cite{liu2026oawam}; ABot-M0.5 extends WAMs to unified mobility-and-manipulation control with action-space disentanglement and train-test consistency mechanisms~\cite{chen2026abot}; and VT-WAM incorporates visual-tactile prediction for contact-rich manipulation~\cite{tian2026vtwam}. These systems illustrate the rapid expansion of WAMs from visual manipulation to richer embodied settings. As WAMs become more capable and more deeply integrated into robot control stacks, understanding their security properties becomes increasingly urgent and meaningful.

Prior WAM work usually treats future prediction as a beneficial representation, planning interface, or safety signal~\cite{liu2026jailwam,chen2026attackingtrusted}. A natural hope is that an imagined future can be used to check whether a planned action is physically plausible or safe. Formally, one might imagine a monitor $M$ that inspects the predicted future:
\begin{equation}
    \mathrm{safe} = M(z_{t+1:t+K}, g).
\end{equation}
However, such a monitor only observes the imagined future, not necessarily the
alignment between the future and the action that will be executed. This leaves
open a WAM-specific security question: can an adversary perturb the observation
so that the action changes substantially while the imagined future remains
apparently plausible? BadWAM formalizes and evaluates this gap in
Section~\ref{sec:method}.

\subsection{Attacks against Embodied AI Systems}

Adversarial examples were first studied extensively in image classification, where small input perturbations can cause large changes in model predictions~\cite{szegedy2014intriguing,goodfellow2015explaining,madry2018towards}. Subsequent work extended these attacks to black-box and query-limited settings, including gradient-free and score-based estimation methods~\cite{chen2017zoo,ilyas2018blackbox,uesato2018adversarial}. Physical-world attacks further showed that adversarial patterns can remain effective under real-world transformations, for example through adversarial patches or robust physical perturbations~\cite{brown2017adversarialpatch,eykholt2018robust}. BadWAM builds on this broader adversarial robustness literature but targets a different object: not a classifier label or static perception output, but the closed-loop action behavior of a WAM.

Attacks on embodied AI systems are particularly concerning because model failures can translate into physical consequences. Prior work has studied attacks on perception modules, reinforcement-learning agents, and embodied policies, including perturbations that alter observations, timing, or environmental cues~\cite{lin2017tactics,gleave2019adversarial}. These attacks typically target reactive decision-making. For a reactive policy $\pi_\theta$, an observation attack can be written as
\begin{equation}
    \max_{\|\delta\|_\infty \le \epsilon}
    D\!\left(
        \pi_\theta(o_t + \delta, g),
        \pi_\theta(o_t, g)
    \right),
\end{equation}
or, in closed-loop evaluation, as minimizing task reward or success. BadWAM naturally generalizes this view to predictive-action models whose outputs include both actions and imagined futures. The relevant question is no longer only whether the action changes, but whether the action changes while the imagination remains plausible.

Recent work has begun to examine attacks and safety risks for world models and WAMs. BadWorld attacks visual world models by perturbing context images to degrade future rollouts under uncertain controls~\cite{shen2026badworld}. Physical-conditioned attacks study how perturbing condition channels in generative world models can induce semantic or decision-level distortions~\cite{zhang2026physcondwma}. JailWAM proposes a safety evaluation framework for WAM jailbreaks in robot control, focusing on unsafe or destructive behaviors represented through visual trajectories~\cite{liu2026jailwam}. These works highlight that predictive models in embodied systems introduce new attack surfaces beyond conventional action policies.

Closest to our setting is recent work on oracle-level integrity attacks against imagine-then-act world models~\cite{chen2026attackingtrusted}. That work studies VLA-style and world-model-based systems, where an imagined future is consumed by downstream oracles such as safety gates, MPC planners, or imagine-then-check verifiers. It shows that corrupting the trusted imagination can compromise systems that rely on future predictions. BadWAM studies a complementary failure mode. Rather than attacking the imagined future as the primary object, we target the alignment between imagination and action in deployed WAM controllers. We show that action and imagination can be adversarially desynchronized: the predicted future may remain close to the clean prediction while the executed action changes enough to cause task failure.

This distinction is central to our threat model. If a safety monitor trusts plausible imagined futures, then a purely imagination-based check may miss an attack that preserves future predictions while hijacking actions. BadWAM therefore treats the alignment between action and imagination as the security-critical object. Our action-only adversarial attack prioritizes task disruption, while our imagination-preserving adversarial attack explicitly studies the trade-off between attack strength and stealthiness.

\section{Threat Model}
\label{sec:threat-model}

We study inference-time attacks against deployed WAM-based robot policies. Following the notation introduced above, the robot observes $o_t$ and receives an instruction or goal $g$ at each replanning step. The WAM then outputs an action chunk $a_{t:t+H-1}$ and, depending on the model interface, may also expose an imagined future in latent form $z_{t+1:t+K}$ or decoded video form $v_{t+1:t+K}$. The robot executes part of the predicted action chunk, observes the environment again, and repeats this process in closed loop.

\noindent\textbf{Adversarial capability.}
The adversary can perturb the visual observation before it is processed by the WAM. We model a digital observation-channel adversary that can intercept the current camera tensor, submit black-box queries, and replace the tensor delivered to the policy. We do not claim that the same perturbations are directly realizable as physical-world patterns. We write the perturbed observation as
\begin{equation}
    \tilde{o}_t = \mathrm{clip}(o_t + \delta_t),
    \quad
    \|\delta_t\|_\infty \le \epsilon .
\end{equation}
The perturbation is bounded so that the attacked observation remains visually close to the clean observation. For multi-camera inputs, we treat $o_t$ as the concatenation of all camera observations and apply the same norm constraint to the full visual input. The adversary does not modify the language instruction $g$, the robot state, the environment dynamics, the model parameters, or the training data.

We consider two access levels. In the \emph{action-only} setting, the adversary can query the WAM and observe only its output action chunk. This captures black-box access to a deployed policy API or robot inference stack. In the \emph{imagination-visible} setting, the adversary can additionally observe the model's imagined future, either as latent rollouts or decoded future video. This captures systems where future prediction is exposed for planning, debugging, monitoring, or imagine-then-check safety verification~\cite{chen2026attackingtrusted,Zhao-RSS-24}. In both settings, the adversary does not require gradients or white-box access to model weights.

\noindent\textbf{Adversarial objective.}
The adversary's primary goal is untargeted task failure: the robot should fail to complete the instructed task under closed-loop execution. The attack is successful if a perturbation causes the robot to execute a sequence of actions that reduces task success, even though the instruction and environment remain unchanged. Since task reward and success labels are unavailable during optimization, BadWAM uses clean-attacked action deviation as a queryable surrogate objective.

\begin{figure}[t]
    \centering
    \includegraphics[width=\columnwidth]{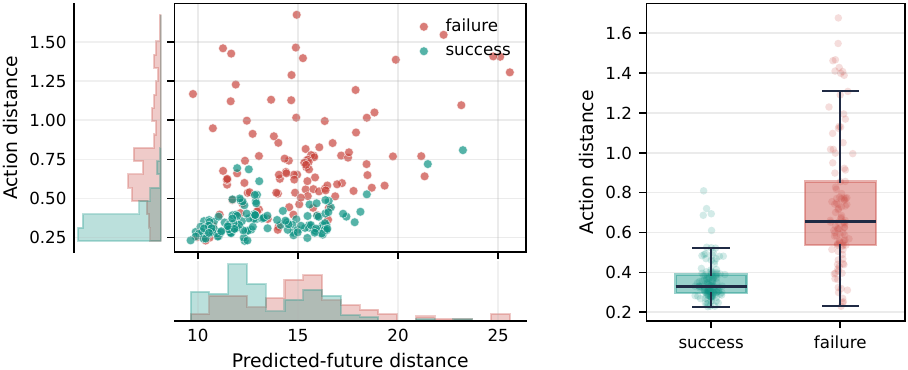}
    \caption{
    Empirical motivation for world-action adversarial attacks. Failed episodes
    tend to have larger action shifts, while predicted-future shifts overlap
    across successful and failed executions. This motivates attacking the
    alignment between action and imagination.
    }
    \label{fig:method-desync-evidence}
    \vspace{-4mm}
\end{figure}
BadWAM additionally studies stealthiness. A non-stealthy attack may simply maximize action disruption. A stealthier attack should degrade execution while keeping the WAM's imagined future close to its clean prediction. This is important because future prediction is often treated as a safety signal: a monitor may inspect the imagined future and accept the action if the future appears plausible~\cite{chen2026attackingtrusted}. BadWAM targets precisely this gap between what the WAM imagines and what it executes.

\noindent\textbf{Non-goals.}
We do not study training-time poisoning, model extraction, reward hacking, prompt injection, or attacks that physically modify the robot's environment. We also do not assume that the attacker can directly command the robot or alter the controller after actions are produced. Our focus is narrower: small bounded observation perturbations that exploit the world-action interface of WAMs at inference time.

\begin{figure}[t]
    \centering
    \includegraphics[width=\columnwidth]{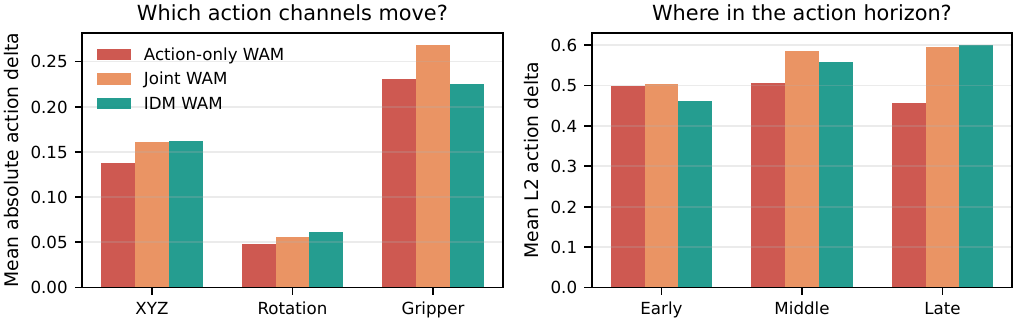
    }
    \caption{
    Action-only adversarial attack produces structured action shifts. Across
    three WAM variants, the attack primarily perturbs continuous action
    channels and specific portions of the action horizon, rather than acting as
    uniform output noise.
    }
    \label{fig:method-action-structure}
    \vspace{-3mm}
\end{figure}

\begin{figure*}[t]
    \centering
    \includegraphics[width=\textwidth]{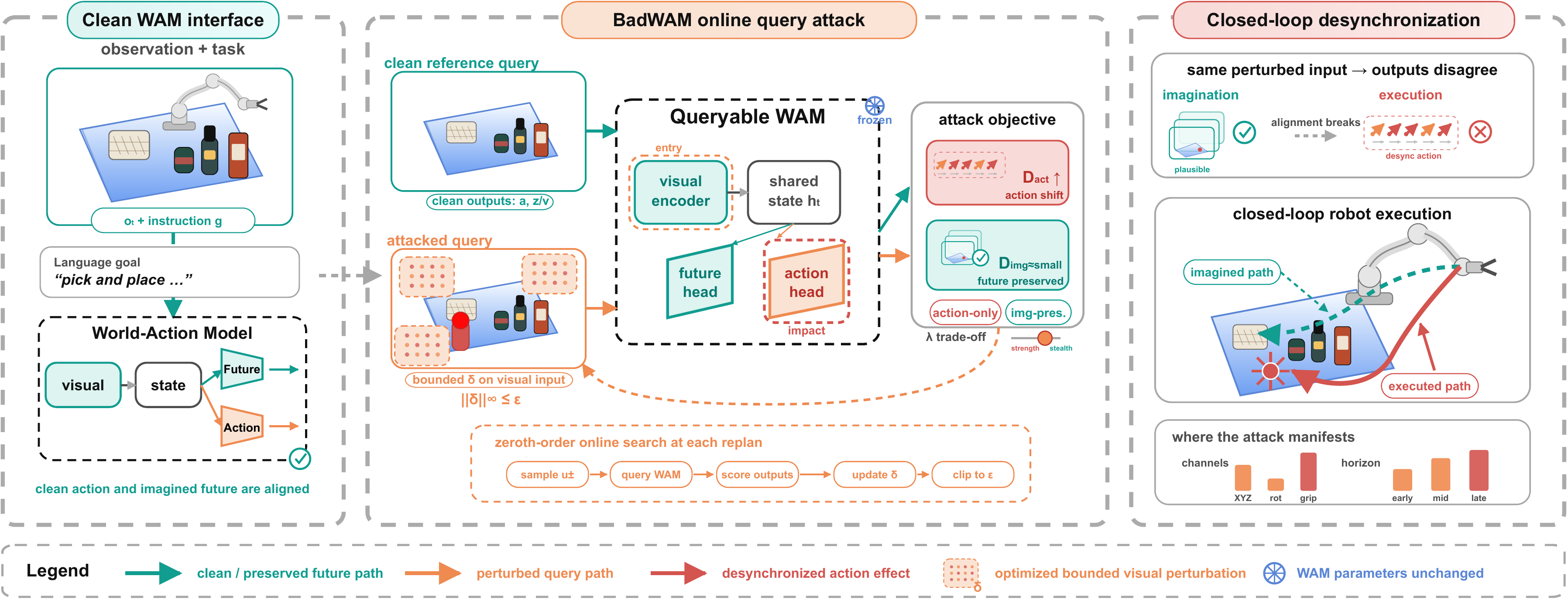}
    \caption{Overview of BadWAM. BadWAM injects a small visual perturbation into model observations and performs query-based online search over a frozen WAM. The optimized perturbation disrupts the action prediction pathway while preserving or minimally altering visual rollout predictions, leading to world-action adversarial attacks during closed-loop execution.
    }
    \label{fig:overall_dia}
    \vspace{-5mm}
\end{figure*}
\section{BadWAM}
\label{sec:method}

BadWAM is a unified framework for modeling and evaluating world-action
drift attacks. As shown in Figure~\ref{fig:overall_dia}, the framework treats a WAM as a queryable
predictive-action system and optimizes bounded visual perturbations at each
replanning step. Its two attack instantiations correspond to two points on the
attack-strength/stealthiness spectrum: an \emph{action-only adversarial
attack} that prioritizes execution failure, and an
\emph{imagination-preserving adversarial attack} that additionally
constrains predicted-future drift.

\subsection{Design Motivation}

BadWAM is motivated by a simple empirical observation: action and imagination are coupled in WAMs, but they are not inseparable. In our closed-loop traces, task failures tend to exhibit larger action shifts than successful episodes, while the corresponding imagination shifts can substantially overlap. This suggests that the security-critical object is not only the action or the imagination alone, but their alignment. Figure~\ref{fig:method-desync-evidence} visualizes this phenomenon under our imagination-preserving attack. This motivation is also consistent with recent diagnostic evidence that WAM action heads can rely on future-predictive latents in a task-phase-dependent manner~\cite{mishra2026temporalratio}. BadWAM asks a complementary security question: whether this coupling can be adversarially utilized.

We therefore formulate BadWAM around two distances. $D_{\mathrm{act}}$ measures
the deviation between clean and attacked action chunks, and $D_{\mathrm{img}}$
measures the drift between clean and attacked imagination. These distances are
interface-level quantities: they do not require a particular WAM architecture,
only access to the corresponding outputs. Conceptually, a stealthy
adversarial attack seeks large action deviation under bounded visual
perturbations while keeping imagination drift below a monitorable threshold:
\begin{equation}
    \max_{\|\delta\|_\infty \le \epsilon}
    D_{\mathrm{act}}(a^\delta, a)
    \quad
    \mathrm{s.t.}\;
    D_{\mathrm{img}}(z^\delta, z) \le \tau_{\mathrm{img}} .
    \label{eq:desync-constrained}
\end{equation}
Equation~\ref{eq:desync-constrained} defines the core BadWAM attack surface.
The two attacks below instantiate different relaxations of this objective.

\subsection{Action-Only Adversarial Attack}

The action-only attack captures the high-strength end of BadWAM. 
It treats the
WAM as a queryable controller and focuses purely on execution disruption. Rather
than enforcing any constraint on the imagined future, it searches for a bounded
visual perturbation that maximizes the deviation between clean and attacked
action chunks:

\begin{equation}
    \delta_t^\star =
    \arg\max_{\|\delta_t\|_\infty \le \epsilon}
    D_{\mathrm{act}}
    \left(
        a_{t:t+H-1}^{\delta},
        a_{t:t+H-1}
    \right).
    \label{eq:action-only}
\end{equation}

The clean action chunk is used only as a reference; the attacker does not need
ground-truth expert actions or task success labels during optimization. This
makes the attack applicable to deployed policy interfaces where the observable
output is the action command sent to the robot controller.

Although simple, this objective is already highly disruptive. In our closed-loop experiments, the action-only attack reduces one WAM variant from 96.5\% to 43.1\% task success. We therefore use it as the high-strength endpoint of BadWAM.
At the same time, the attack is not merely unstructured output corruption: the induced shifts are unevenly distributed across both action channels and temporal horizon segments. For example, Figure~\ref{fig:method-action-structure} shows that the perturbations concentrate on task-relevant channels such as translation and gripper commands, while their temporal concentration varies across WAM variants. This motivates logging channel-level and horizon-level action statistics in our evaluation.

\begin{figure*}[t]
    \centering

    \begin{subfigure}{0.48\linewidth}
        \centering
        \includegraphics[width=\linewidth]{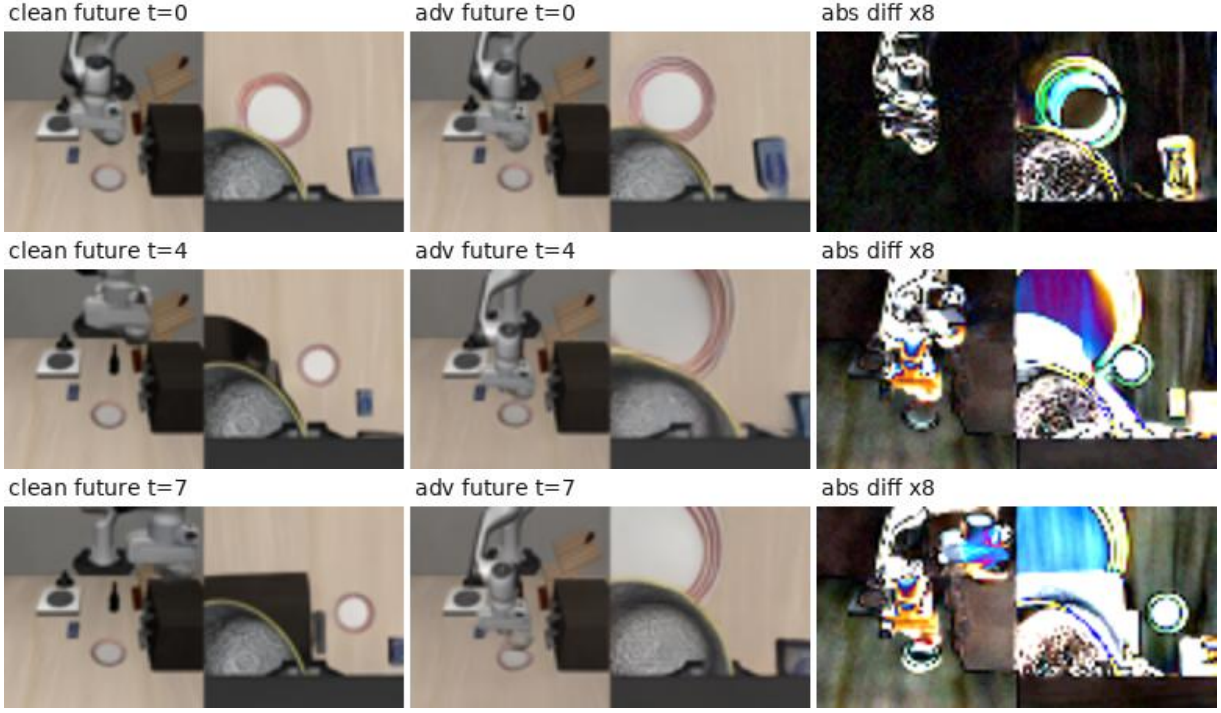}
        \caption{Imagined future without preservation.}
        \label{fig:clean}
    \end{subfigure}
    \hfill
    \begin{subfigure}{0.48\linewidth}
        \centering
        \includegraphics[width=\linewidth]{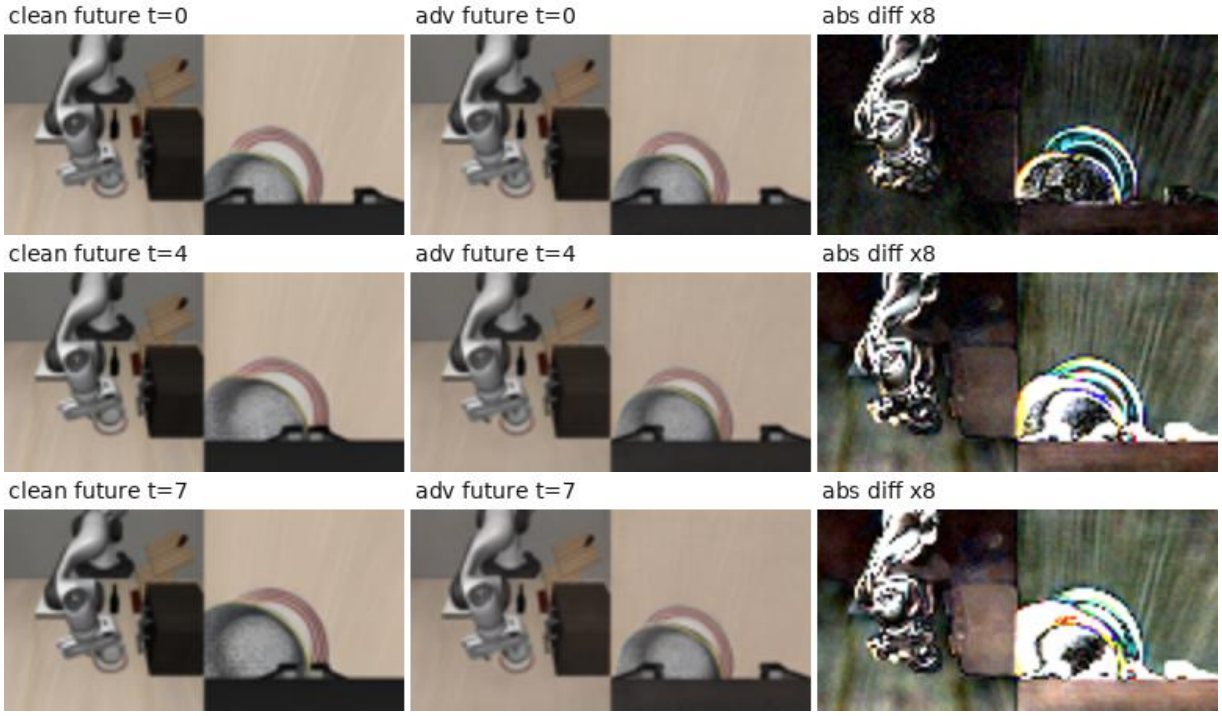}
        \caption{Imagined future with preservation.}
        \label{fig:adv}
    \end{subfigure}
    \vspace{-3mm}
\caption{
Qualitative illustration of the imagination-preserving objective on IDM WAM.
Each panel compares predicted futures under clean and adversarial observations
at selected future steps.
\texttt{abs diff} denotes the absolute pixel difference between clean and
adversarial predictions, amplified by $8\times$ for visibility.
Both variants induce action-space failure, but the preservation term keeps the
adversarial future more consistent with the clean imagination.
}
\vspace{-4mm}
    \label{fig:future_preserving_case}
\end{figure*}

\subsection{Imagination-Preserving Adversarial Attack}

The action-only objective provides no control over what happens to the imagined
future. For WAM deployments that expose imagined futures to planners, monitors,
or imagine-then-check safety gates, this is insufficient: a stealthier attack
should degrade execution while preserving the model's clean imagination.
BadWAM therefore instantiates an imagination-preserving attack by solving the
Lagrangian relaxation of Equation~\ref{eq:desync-constrained}:

\begin{equation}
\begin{aligned}
    \delta_t^\star
    =
    \arg\max_{\|\delta_t\|_\infty \le \epsilon}
    \;&
    D_{\mathrm{act}}
    (a_{t:t+H-1}^{\delta}, a_{t:t+H-1}) \\
    &-
    \lambda
    D_{\mathrm{img}}
    (z_{t+1:t+K}^{\delta}, z_{t+1:t+K}) .
\end{aligned}
\label{eq:imagination-preserving}
\end{equation}

The coefficient $\lambda$ controls the trade-off between attack-strength and stealthiness.
When $\lambda=0$, the objective reduces to pure action disruption. As
$\lambda$ increases, the optimizer is discouraged from changing the imagined
future, making the attack more stealthy but potentially less damaging.

When decoded future videos are available, we instantiate the imagination
distance as an average frame-level distance:
\begin{equation}
    D_{\mathrm{img}}
    (v_{t+1:t+K}^{\delta}, v_{t+1:t+K})
    =
    \frac{1}{K}
    \sum_{k=1}^{K}
    d_{\mathrm{frame}}
    (v_{t+k}^{\delta}, v_{t+k}) .
    \label{eq:video-distance}
\end{equation}
The same objective can be applied to latent futures by replacing the
frame-level distance with a latent-space distance. In either case, the attack
targets a WAM-specific failure: the robot may appear to imagine a plausible
future while executing a desynchronized action.

\begin{table*}[t]
    \centering
    \caption{
    Main closed-loop attack results on LIBERO and RoboTwin, with $\epsilon=0.06$ and eight optimization iterations
($B=16$ perturbation queries).
    Gray values indicate the absolute reduction from clean
    task success.
    }
    \label{tab:main-attack-results}
    \begin{tabular}{lcccccc}
        \toprule
        \hline
        & \multicolumn{3}{c}{LIBERO (\%)} 
        & \multicolumn{3}{c}{RoboTwin (\%)} \\
        \cmidrule(lr){2-4} \cmidrule(lr){5-7}
        Model 
        & Clean & Action-only & Img-pres.
        & Clean & Action-only & Img-pres. \\
        \midrule
        Action-only WAM 
        & 96.5
        & 43.1 {\scriptsize\color{gray}($\downarrow$53.4)}
        & --   
        & 92.1
        & 84.4 {\scriptsize\color{gray}($\downarrow$7.7)}
        & --  \\
        
        Joint WAM       
        & 98.1
        & 61.5 {\scriptsize\color{gray}($\downarrow$36.6)}
        & 63.0 {\scriptsize\color{gray}($\downarrow$35.1)}
        & 93.3 
        & 86.0 {\scriptsize\color{gray}($\downarrow$7.3)}
        & 87.0 {\scriptsize\color{gray}($\downarrow$6.3)}
        \\
        
        IDM WAM         
        & 98.4
        & 66.1 {\scriptsize\color{gray}($\downarrow$32.3)}
        & 67.0 {\scriptsize\color{gray}($\downarrow$31.4)}
        & 92.9
        & 75.4 {\scriptsize\color{gray}($\downarrow$17.5)}
        & 76.5 {\scriptsize\color{gray}($\downarrow$16.4)}
        \\
        \hline
        \bottomrule
    \end{tabular}
    \vspace{-5mm}
\end{table*}

\subsection{Query-Based Online Optimization}

BadWAM does not backpropagate through the WAM or access model parameters. At
each replanning step, the attacker treats the WAM as a queryable input-output
system and optimizes the attack objective with zeroth-order finite-difference
queries. Concretely, for a current perturbation $\delta_t$, the attacker samples
random directions $u_i$ and evaluates the scalar objective under positively and
negatively perturbed observations. This gives the estimator
\begin{equation}
    \widehat{\nabla} J(\delta_t)
    =
    \frac{1}{m}
    \sum_{i=1}^{m}
    \frac{
        J(\delta_t + c u_i) - J(\delta_t - c u_i)
    }{2c}
    u_i ,
    \label{eq:zo-estimator}
\end{equation}
where $J$ denotes either the action-only objective or the
imagination-preserving objective, $c$ is the finite-difference radius,
and $m$ is the number of sampled directions. The perturbation is then updated
and projected back into the feasible $\ell_\infty$ ball:
\begin{equation}
    \delta_t
    \leftarrow
    \Pi_{[-\epsilon,\epsilon]}
    \left(
        \delta_t + \eta \widehat{\nabla} J(\delta_t)
    \right).
    \label{eq:query-update}
\end{equation}
This optimization is zeroth-order because it only uses objective values computed
from WAM outputs; it does not require gradients, model weights, or training
data. In our implementation, the random directions follow a simultaneous
perturbation strategy~\cite{spall1992multivariate,uesato2018adversarial}. The
same optimizer supports both BadWAM instantiations by changing only the scalar
objective $J$: the action-only attack scores action deviation, while the
imagination-preserving attack scores action deviation together with
future-prediction consistency.

For each replanning step, BadWAM keeps the best perturbation found within the
query budget. The final adversarial observation is then used to produce the
action chunk executed by the robot. This online design avoids training a
separate attack model and naturally adapts to the current observation,
instruction, and robot state.

\noindent\textbf{Qualitative Example}
Figure~\ref{fig:future_preserving_case} qualitatively illustrates the role of the preservation term. Both variants can induce action failure, but adding preservation keeps the predicted future closer to the clean imagination. This is precisely the WAM-specific failure mode BadWAM targets: the model's imagination remains plausible while the executed action is desynchronized.

\subsection{Closed-Loop Execution and Measurements}

BadWAM is evaluated under closed-loop execution. The attack is recomputed at each replanning step, the WAM outputs an attacked action chunk, and the robot executes the selected actions before the next replan. This setting is more faithful than single-step action comparison because small action shifts may compound over time and cause task failure only after multiple replans.
We report three groups of measurements. First, we measure task-level attack effectiveness using closed-loop success rate and induced failures relative to clean execution. Second, we measure action disruption using action distance, action-channel statistics, and horizon-level action shifts. Third, for imagination-preserving attacks, we measure stealthiness using predicted-future distance and adversarial score. Together, these metrics separate two questions: \textit{whether the robot fails, and whether the failure is visible through the WAM's imagined future.}

\noindent\textbf{Positioning. }We \textit{do not} execute BadWAM on physical robots because the evaluation deliberately searches for task-failing controls whose consequences cannot be bounded in advance. Even small observation perturbations may accumulate into abrupt contacts, collisions, dropped objects, or motions outside the safe workspace, posing risks to hardware and nearby operators. We therefore conduct the closed-loop evaluation in simulation, which enables repeatable initial conditions, matched clean and attacked rollouts, and systematic full-sweep testing without avoidable physical harm. Real-robot validation would require an independent safety layer, conservative motion and force limits, and hardware-specific supervision, which we leave to future work.

\section{Evaluation}
\label{sec:evaluation}

We organize the evaluation around six research questions:

\begin{figure}[t]
    \centering
    \includegraphics[width=\columnwidth]{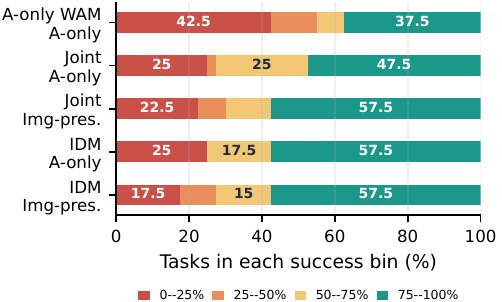}
    \caption{
    Task-level failure profile under attack on LIBERO. This figure groups the 40 LIBERO tasks according to their attacked closed-loop success rates, using 20 trials per task. Lower bins indicate tasks that are consistently broken rather than only mildly degraded.
    }
    \label{fig:exp-main-results}
    \vspace{-4mm}
\end{figure}

\begin{itemize}
    \item[\textbf{RQ1}] \textbf{Attack effectiveness.}
    How much can bounded observation perturbations reduce closed-loop WAM reliability?

    \item[\textbf{RQ2}] \textbf{Systematicity.}
    Are the failures concentrated in a few tasks, or do they persist across suites and repeated trials?

    \item[\textbf{RQ3}] \textbf{Failure mechanism.}
    What do BadWAM failures look like during execution, and how does the query-based search produce them?

    \item[\textbf{RQ4}] \textbf{Stealth under comparable attack effectiveness.}
    At comparable attack effectiveness, what does imagination preservation buy?

    \item[\textbf{RQ5}] \textbf{Sensitivity and efficiency.}
    How do future preservation, perturbation size, and query budget trade off?

    \item[\textbf{RQ6}] \textbf{Transferability.}
    Do the perturbations transfer across WAM variants?
\end{itemize}

\subsection{Experimental Setup}
\label{sec:eval-setup}

\noindent\textbf{Benchmarks.}
Our main experiments use two language-conditioned robot manipulation benchmarks:
LIBERO~\cite{liu2023libero} and RoboTwin~\cite{chen2025robotwin}.
Unless otherwise stated, each LIBERO main-result cell uses all 40 tasks with 20 trials per task, and each RoboTwin main-result cell uses all 50 tasks with 20 trials per task. Clean and attacked conditions use the same task definitions and evaluation seed protocol. Studies using a different scope report it explicitly in their caption.

\begin{figure}[t]
    \centering
    \includegraphics[width=\columnwidth]{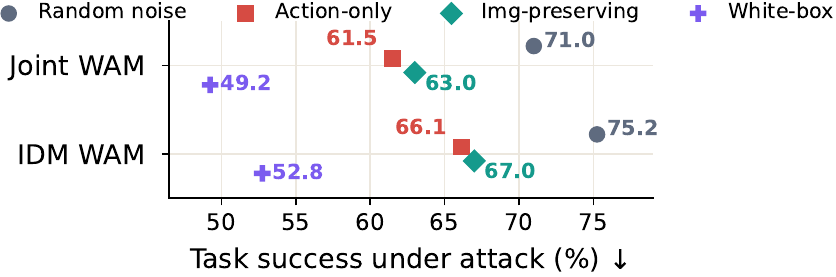}
    \caption{
    Comparison with random perturbations on full LIBERO benchmark.
    All methods use the same $\ell_\infty$ budget.
    Lower task success indicates a stronger attack.
    }
    \label{fig:random-baseline}
    \vspace{-5mm}
\end{figure}
\noindent\textbf{WAM policies.}
We evaluate three WAM-style controllers that expose different interfaces~\cite{yuan2026fastwam,shen2026world}.
The action-only WAM maps the current observation and instruction directly to an action sequence.
The joint WAM predicts future visual states and actions jointly.
The IDM WAM first constructs a future-imagination representation and then decodes actions from it.
This model set separates action-only policies from policies whose imagined futures are also query-visible.
All evaluations are closed-loop: the policy is repeatedly queried during execution, and BadWAM may perturb the current observation at each replan. For the action-only variant on LIBERO and RoboTwin, we use the officially released checkpoints. For the other two variants on LIBERO and RoboTwin, we train them on 8 H100 GPUs with the default settings~\cite{yuan2026fastwam}. The details for the training process are reported in Appendix~\ref{training_details}.

\begin{figure*}[t]
    \centering
    \includegraphics[width=\textwidth]{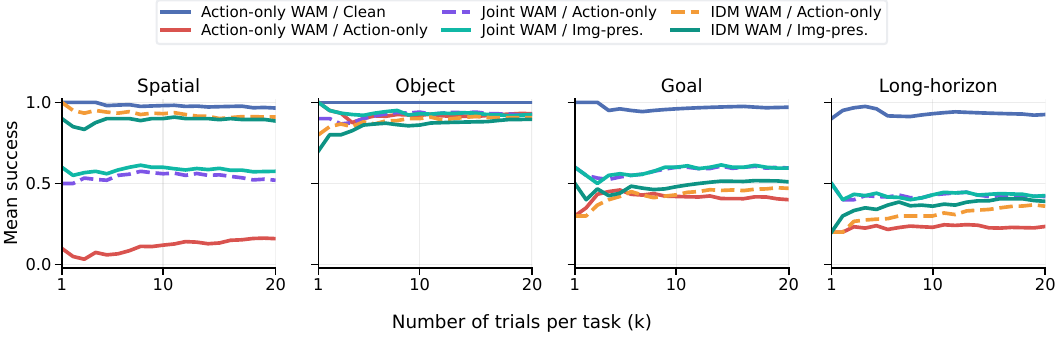}
    \vspace{-5mm}
    \caption{
    Mean pass@$k$ across trials on different LIBERO suites.
    BadWAM lowers success throughout the trial budget instead of only causing isolated unlucky failures.
    }
    \label{fig:exp-pass-at-k}
\vspace{-5mm}
\end{figure*}

\noindent\textbf{Attacks.}
We evaluate two objectives.
The \emph{action-only adversarial attack} maximizes the shift in the predicted action sequence under an $\ell_\infty$-bounded observation perturbation.
It requires only action outputs and therefore applies to all three policies.
The \emph{imagination-preserving adversarial attack} additionally penalizes changes in the predicted future, so it applies to WAMs whose future predictions are query-visible.
Unless otherwise stated, both attacks perturb the full image with $\epsilon=0.06$ on FastWAM's normalized $[-1,1]$ image tensor (approximately $8/255$ on the original RGB scale, away from clipping).
The default attack uses eight paired finite-difference updates per replan, corresponding to a perturbation-query budget of $B=16$ plus one clean reference query, for 17 WAM forward queries in total. Here $B$ counts perturbation evaluations, while the reported total query count also includes the clean reference. Ablation studies can be found in Section~\ref{sec:eval-ablations}.

\noindent\textbf{Metrics.}
Our primary metric is closed-loop task success rate; lower success under attack indicates a stronger attack.
For a benchmark with $T$ tasks and $N_i$ evaluation episodes for task $i$, we compute
\begin{equation}
    \mathrm{Success}
    =
    \frac{1}{T}
    \sum_{i=1}^{T}
    \frac{1}{N_i}
    \sum_{j=1}^{N_i}
    \mathbb{I}\!\left[s_{i,j}=1\right],
\end{equation}
where $s_{i,j}$ indicates whether episode $j$ of task $i$ succeeds.
We also report the induced success drop.
For mechanism analysis, we use action distance $D_{\mathrm{act}}$, predicted-future distance $D_{\mathrm{img}}$, and the normalized decoupling score from Section~\ref{sec:method}.
For repeated-trial analysis, we report pass@$k$: for each task, pass@$k$ here is the fraction of successful executions among its first $k$ trials, averaged across tasks.

\subsection{BadWAM Reliably Induces Task Failures}
\label{sec:eval-main-result}

To answer \textbf{RQ1}, Table~\ref{tab:main-attack-results} reports the main results on LIBERO and RoboTwin. All clean WAM policies achieve strong task performance. On LIBERO, the action-only, joint, and IDM WAMs obtain clean success rates of 96.5\%, 98.1\%, and 98.4\%, respectively. The action-only attack reduces these rates to 43.1\%, 61.5\%, and 66.1\%. On the joint and IDM WAMs, the imagination-preserving attack achieves success rates of 63.0\% and 67.0\%. The attack also generalizes to RoboTwin. From clean success rates of 92.1\%, 93.3\%, and 92.9\%, the action-only attack reduces performance to 84.4\%, 86.0\%, and 75.4\%, respectively. The imagination-preserving attack lowers the joint and IDM WAMs to 87.0\% and 76.5\%. These results show that coupling action generation with future prediction does not inherently protect WAMs against small, bounded observation perturbations. Moreover, the imagination-preserving attack remains close in strength to the action-only attack on both datasets, despite explicitly suppressing predicted-future drift. BadWAM can therefore preserve the imagination pathway while still shifting the action pathway sufficiently to cause task failure.

Figure~\ref{fig:exp-main-results} complements the aggregate table with a task-level failure profile. For the action-only WAM, 42.5\% of tasks fall into the 0-25\% success bin under attack. For the joint and IDM WAMs, 17.5-25.0\% of tasks fall into this lowest-success bin, depending on the objective. This distributional view is important for safety: a moderate average can hide tasks that become almost unusable.

\begin{figure*}[t]
    \centering
    \includegraphics[width=\textwidth]{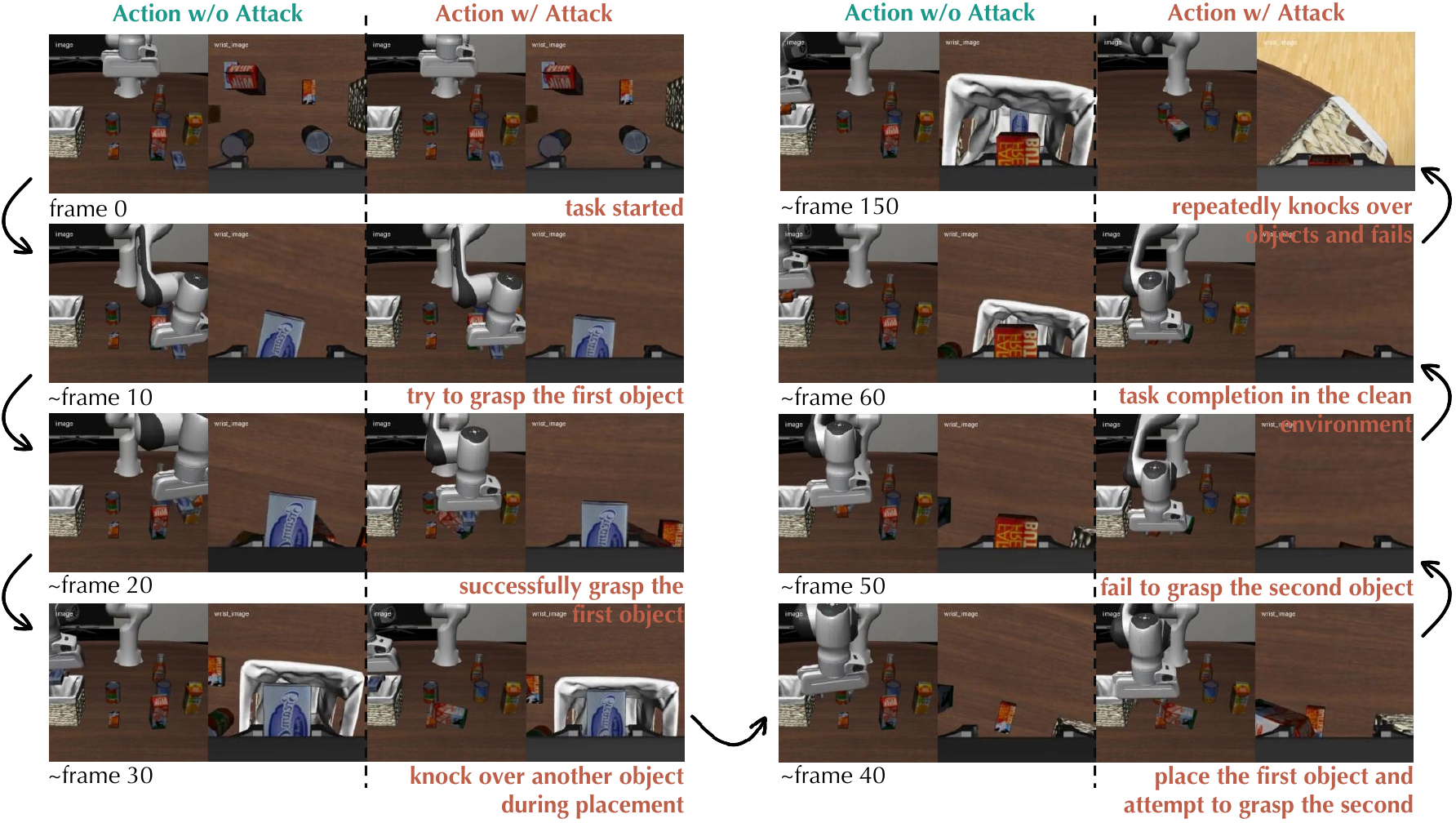}
    \caption{
Qualitative comparison of clean and attacked rollouts on a LIBERO-10 task.
Each row shows synchronized third-person and wrist-camera observations over time.
The clean WAM completes the sequential manipulation task, while the attacked WAM initially behaves plausibly but gradually drifts, knocks over objects, fails to grasp the second object, and eventually fails.
    }
    \label{fig:actual_actions}
    \vspace{-5mm}
\end{figure*}
\noindent\textbf{Sanity check against white-box attacks and random perturbations.}
To verify that BadWAM is not merely exploiting generic visual fragility, we compare it with two references in Figure~\ref{fig:random-baseline}.
First, we use a model-agnostic random uniform perturbation baseline under the same $\ell_\infty$ budget.
Second, we include a stronger white-box reference that backpropagates through the WAM to optimize an $\ell_\infty$-bounded input perturbation with a differentiable action-output surrogate; after the perturbation is found, the model is still evaluated with the standard closed-loop inference procedure.
This white-box setting is outside BadWAM's black-box threat model, but serves as an upper-bound reference for how much stronger attacks can become with gradient access.

\begin{figure*}[t]
    \centering
    \includegraphics[width=\textwidth]{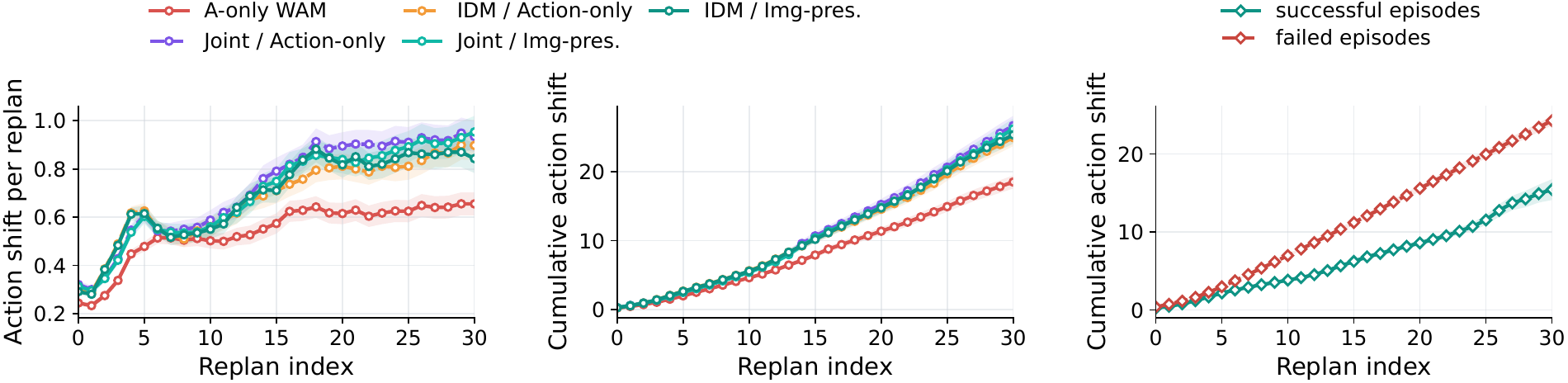}
\caption{
Per-replan action shifts persist across execution and accumulate over time, with failed episodes showing substantially larger cumulative shifts than successful ones.
Action shift is measured between clean and attacked action chunks at each replan; shaded regions denote 95\% confidence intervals.
}
    \label{fig:exp-accumulation}
    \vspace{-3mm}
\end{figure*}
Random perturbations are much weaker than targeted attacks: on Joint WAM and IDM WAM, random noise leaves task success at 71.0\% and 75.2\%, while BadWAM reduces success to 61.5\%/63.0\% and 66.1\%/67.0\% for action-only and imagination-preserving objectives, respectively.
The white-box reference further lowers success to 49.2\% and 52.8\%, showing that stronger access can amplify the vulnerability.
Together, these results suggest that BadWAM failures are caused by targeted world-action attacks rather than arbitrary image corruption, and that our black-box attacks are strong and also leave room below the white-box attack with weaker threat model assumption.

\subsection{Systematic Patterns of Failures}
\label{sec:eval-systematic}

To answer \textbf{RQ2}, we test whether the success drop survives repeated trials. Figure~\ref{fig:exp-pass-at-k} plots pass@$k$ as $k$ grows from 1 to 20. If BadWAM only exploited a few unlucky seeds, the attacked curves would move back toward the clean curves as more trials are averaged. Instead, the separation persists across the full trial. The clean action-only WAM stays close to 1.0, while the attacked action-only WAM remains around 0.40-0.43. The joint and IDM WAMs also remain consistently below their clean references under both attack objectives.

\subsection{How Do BadWAM Failures Emerge?}
\label{sec:eval-mechanism}

We next answer \textbf{RQ3} by examining representative executions and the optimization traces that produce them.
Figure~\ref{fig:actual_actions} shows a representative failure.
The clean WAM completes the sequential manipulation task by grasping and placing the required objects in order.
The attacked WAM, however, does not immediately collapse into random motion.
It initially follows a plausible strategy and interacts with the right region, but its actions gradually drift away from the task intent.
As the rollout progresses, the robot knocks over nearby objects, fails to stably grasp the second object, and eventually fails.

This example highlights why closed-loop evaluation is necessary.
The perturbation need not make the observation obviously nonsensical or cause an obviously catastrophic first action.
Instead, it can introduce structured deviations that remain locally plausible but become physically consequential over the horizon.
Figure~\ref{fig:exp-accumulation} confirms this behavior at the dataset level:
per-replan action shifts persist across execution and accumulate over time.
Failed episodes accumulate substantially larger action shifts than successful ones, suggesting that BadWAM failures often emerge from progressive closed-loop drift rather than a single isolated bad action.

\begin{figure}[t]
    \centering
    \includegraphics[width=\columnwidth]{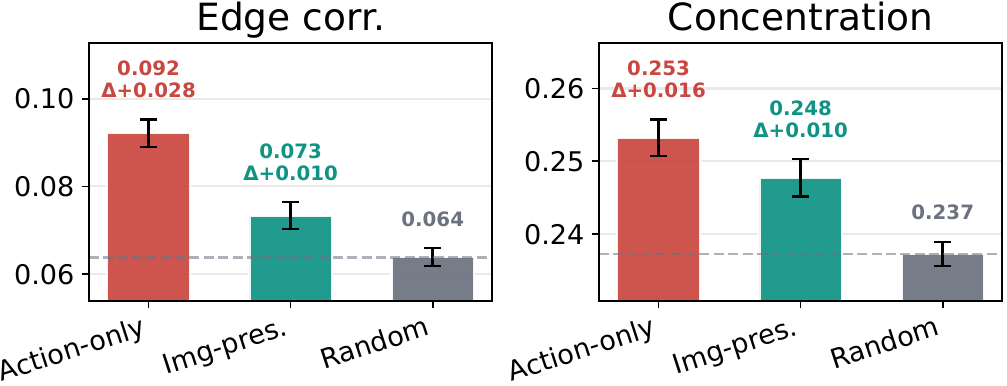}
    \caption{
    Perturbation fingerprints on RoboTwin Joint WAM, using $\ell_\infty$ perturbations with $\epsilon=0.06$ and query budget $B=16$ per replan.
    Compared with random noise, both attacks are more edge-aligned and spatially concentrated.
    Error bars denote 95\% confidence intervals.
    }
    \label{fig:robotwin-perturbation-fingerprint}
    \vspace{-3mm}
\end{figure}

\begin{figure}[t]
    \centering
    \includegraphics[width=\columnwidth]{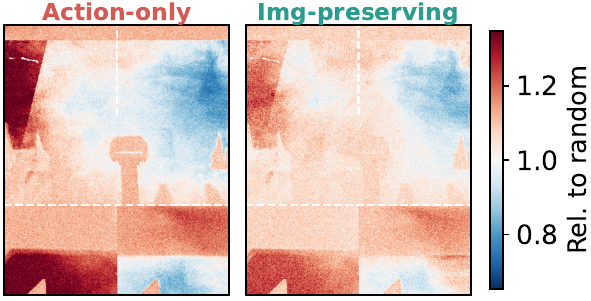}
    \caption{
    Relative perturbation fingerprints on RoboTwin Joint WAM.
    Both BadWAM attacks use full-image $\ell_\infty$ perturbations with $\epsilon=0.06$ and query budget $B=16$ per replan.
    Colors show perturbation strength relative to random noise: red indicates stronger-than-random perturbation energy and blue indicates weaker-than-random energy.
    Dashed lines separate the composed RoboTwin camera views.
    }
    \label{fig:robotwin-perturbation-heatmap}
    \vspace{-3mm}
\end{figure}

\paragraph{Spatial structure of the perturbations.}
We further examine whether BadWAM produces structured input perturbations rather than merely exploiting generic image noise. For each replan, \emph{edge correlation} is the Pearson correlation between per-pixel mean absolute perturbation magnitude and the clean image's gradient magnitude; \emph{concentration} is the Gini coefficient of the spatial perturbation-magnitude map. As shown in Figure~\ref{fig:robotwin-perturbation-fingerprint}, both objectives are more aligned with image edges and more spatially concentrated than random perturbations.

Figure~\ref{fig:robotwin-perturbation-heatmap} further localizes this structure relative to the random baseline. Both objectives redistribute perturbation energy non-uniformly across the composed camera views, amplifying selected regions while suppressing others. The action-only objective produces stronger spatial contrast, whereas the imagination-preserving objective exhibits a milder but related pattern. Because all methods use the same full-image $\ell_\infty$ constraint, these patterns arise from the attack optimization rather than from a hand-designed spatial mask.

\begin{figure*}[t]
    \centering
    \includegraphics[width=\textwidth]{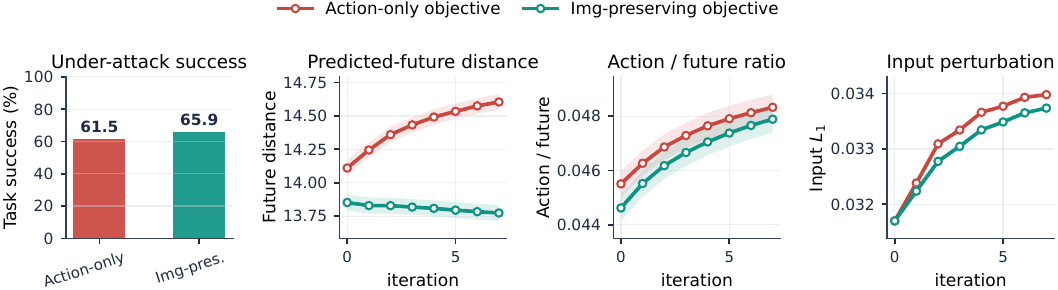}
    \caption{
    Matched-strength stealth trade-off.
    The imagination-preserving objective produces consistently smaller predicted-future shifts under a comparable input perturbation budget, revealing a stealthier failure mode than the action-only objective.
    Shaded regions denote 95\% confidence intervals over replans.
    }
    \label{fig:matched-stealth-dynamics}
    \vspace{-4mm}
\end{figure*}

\begin{figure}[t]
    \centering
    \includegraphics[width=\columnwidth]{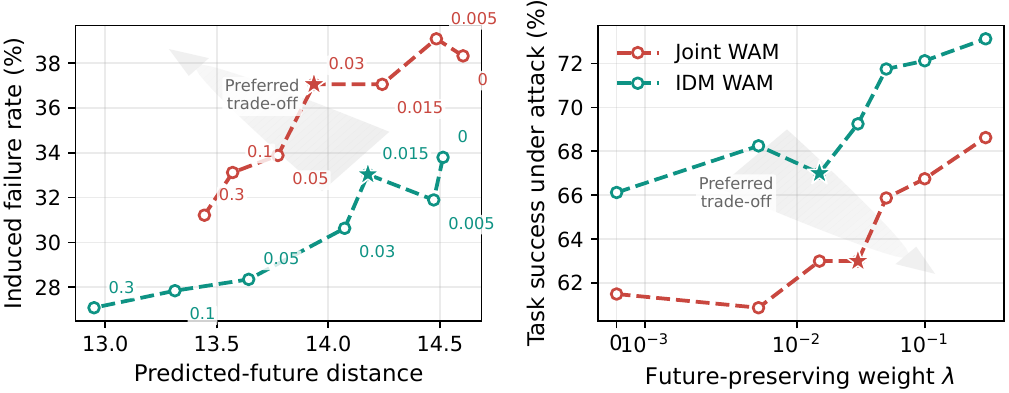}
    \caption{Ablation on future-preserving weight $\lambda$.
    Moderate future preservation can improve the action-future tradeoff, while excessive preservation weakens action manipulation.
    }
    \label{fig:exp-future-weight}
    \vspace{-6mm}
\end{figure}

\begin{figure}[t]
    \centering
    \includegraphics[width=\columnwidth]{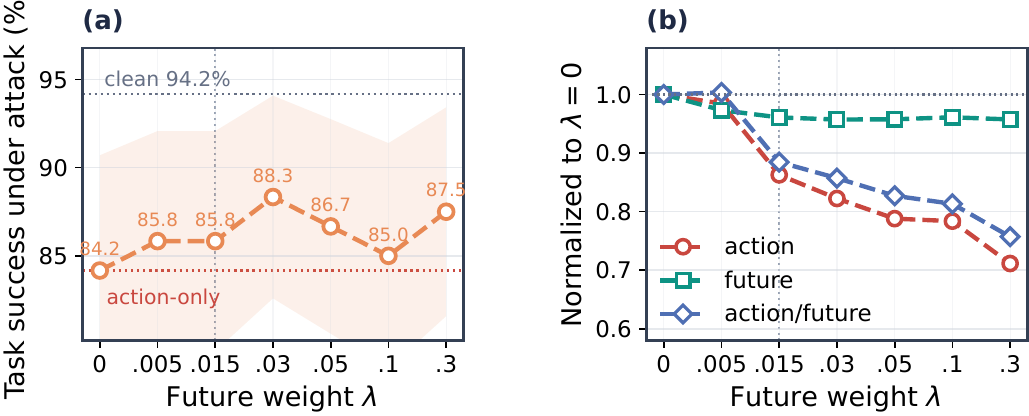}
    \caption{
Effect of the future-preserving weight on the Joint WAM in RoboTwin.
    \textbf{(Left)} Task success under attack, where lower success indicates a stronger attack; dotted horizontal lines mark the clean baseline and the action-only objective at $\lambda=0$.
    \textbf{(Right)} Action shift, future drift, and their ratio, each normalized to $\lambda=0$.
    Increasing $\lambda$ generally reduces action shift while keeping future drift relatively stable, revealing the trade-off between attack-strength and stealthiness.
    }
    \label{fig:robotwin-future-weight}
    \vspace{-6mm}
\end{figure}

\subsection{Imagination Preservation at Comparable Attack Effectiveness}
\label{sec:eval-stealth}
We then answer \textbf{RQ4} by comparing the two objectives under the same perturbation bound, query budget, and closed-loop protocol, while selecting operating points with comparable task-level effectiveness. Figure~\ref{fig:matched-stealth-dynamics} isolates this trade-off under matched attack resources. The two objectives use the same perturbation bound, query budget, and closed-loop protocol. Their under-attack success rates are close, but the imagination-preserving objective reduces average predicted-future distance from 14.01 to 13.04 while keeping the input perturbation budget essentially matched. Across all the LIBERO tasks in this run, its predicted-future distance is lower on 39 tasks. This supports the core BadWAM claim: a WAM can be pushed toward failure while its imagined future remains closer to the clean prediction than a purely action-only objective would allow.

These dynamics are also useful diagnostically.
If action distance rises together with a large future-distance increase, the attack becomes less stealthy.
If future distance stays low but action distance does not improve, the objective is over-regularized.
The default configuration lies between these regimes: it induces a damaging action shift while retaining a comparatively stable predicted future.

\subsection{Ablations: Future Preservation, Perturbation Budget, and Runtime}
\label{sec:eval-ablations}

To answer \textbf{RQ5}, we ablate the future-preserving weight, perturbation budget, and query budget. Figure~\ref{fig:exp-future-weight} on LIBERO shows that increasing $\lambda$ consistently reduces predicted-future distance, but eventually weakens the attack. The induced-failure rate is the fraction of matched clean-success episodes that fail under attack. The preferred trade-offs occur at $\lambda=0.03$ for the Joint WAM and $\lambda=0.015$ for the IDM WAM, where future drift is reduced with only a modest increase in attacked task success. The stars mark representative nondominated points when
higher matched induced failure and lower $D_{\mathrm{img}}$
are jointly preferred. Larger weights further preserve imagination but noticeably reduce action disruption, confirming that $\lambda$ controls the trade-off between stealthiness and attack strength.

We repeat the future-weight ablation on the RoboTwin Joint WAM to test whether the trade-off also appears beyond LIBERO. As shown in Figure~\ref{fig:robotwin-future-weight}, evaluated on a 12-task RoboTwin subset with 10 episodes per task, the default $\lambda=0.015$ achieves 85.8\% task success
under attack, compared with 94.2\% for clean execution
and 84.2\% for the action-only objective. Relative to $\lambda=0$, its action and future distances decrease to 0.86 and 0.96, respectively. As $\lambda$ increases further, the action shift decreases steadily, reaching 0.71 at $\lambda=0.3$, while future drift remains near 0.96 and closed-loop success varies between 84.2\% and 88.3\%. Thus, the action-imagination trade-off also appears on RoboTwin but with different patterns.

\begin{figure*}[t]
    \centering
    \includegraphics[width=\textwidth]{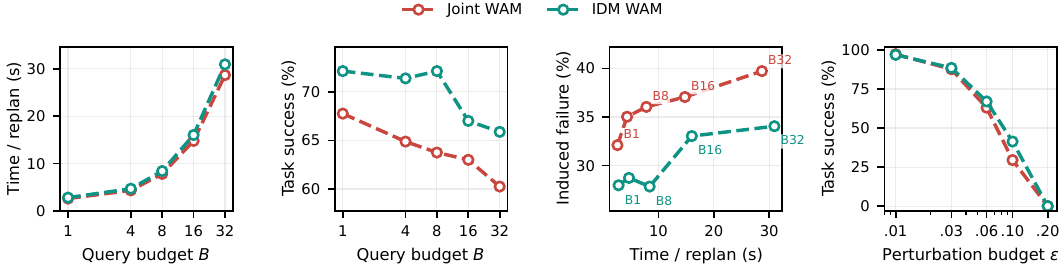}
    \caption{
    Efficiency and budget sensitivity of BadWAM.
    Increasing the query budget $B$ raises the per-replan optimization cost, but also gives the optimizer more opportunity to find stronger perturbations.
    The perturbation-budget study further shows that larger $\epsilon$ consistently improves attack effectiveness.
    Together, the curves expose the practical tradeoff among attack strength, stealthiness, and runtime.
    }
    \label{fig:exp-efficient}
    \vspace{-3mm}
\end{figure*}

Figure~\ref{fig:exp-efficient} studies the budget sensitivity of the imagination-preserving attack on the full LIBERO benchmark. Increasing $\epsilon$ has the strongest effect: success falls from 97.3\% and 96.9\% at $\epsilon=0.01$ to 63.0\% and 67.0\% at $\epsilon=0.06$ for the Joint and IDM WAMs, respectively, and reaches 0.0\% for both at $\epsilon=0.20$. A larger query budget provides moderate gains in attack strength, but incurs a near-linear increase in runtime.

The query-budget ablation shows a clear runtime-effectiveness trade-off. Increasing $B$ from 1 to 32 raises the mean per-replan optimization time from 2.59\,s to 28.68\,s for the Joint WAM and from 2.79\,s to 30.94\,s for the IDM WAM, while reducing task success from 67.8\% to 60.3\% and from 72.1\% to 65.9\%, respectively. Most of this improvement is achieved by $B=16$; doubling the budget to 32 yields only modest additional reductions while nearly doubling runtime. We therefore use the moderate default budget of $B=16$, corresponding to eight optimization iterations. These prototype costs include extensive statistics collection and should not be interpreted as optimized real-time performance.

\begin{table*}[t]
\centering
\caption{
Transferability of BadWAM attacks across WAM variants on LIBERO.
The perturbation is optimized on the source WAM and evaluated on the
target WAM.
``Direct on target'' denotes the same attack objective optimized and
evaluated directly on the target WAM.
Gray values indicate the absolute reduction from target clean task success. $D_{\mathrm{img}}^{\mathrm{direct}}$ and $D_{\mathrm{img}}^{\mathrm{transfer}}$ denote the mean predicted-future distances under direct and transferred attacks, respectively.
}
\label{tab:transferability-results}
\resizebox{0.95\textwidth}{!}{%
\begin{tabular}{ll|cc|c@{\hspace{3pt}}c@{\hspace{3pt}}c|c@{\hspace{3pt}}c@{\hspace{3pt}}c}
    \toprule
    \hline
    Attack objective
    & Source $\rightarrow$ Target
    & Source clean (\%)
    & Target clean (\%)
    & Direct on target (\%)
    &
    & Transferred attack (\%)
    & $D_{\mathrm{img}}^{\mathrm{direct}}$
    &
    & $D_{\mathrm{img}}^{\mathrm{transfer}}$ \\
    \midrule

    Action-only
    & Action-only WAM $\rightarrow$ Joint WAM
    & 96.5
    & 98.1
    & 61.5 {\scriptsize\color{gray}($-36.6\%$)}
    & $\longrightarrow$
    & 64.1 {\scriptsize\color{gray}($-34.0\%$)}
    & --
    &
    & -- \\

    Action-only
    & Joint WAM $\rightarrow$ IDM WAM
    & 98.1
    & 98.4
    & 66.1 {\scriptsize\color{gray}($-32.3\%$)}
    & $\longrightarrow$
    & 70.8 {\scriptsize\color{gray}($-27.6\%$)}
    & --
    &
    & -- \\

    Action-only
    & IDM WAM $\rightarrow$ Joint WAM
    & 98.4
    & 98.1
    & 61.5 {\scriptsize\color{gray}($-36.6\%$)}
    & $\longrightarrow$
    & 65.4 {\scriptsize\color{gray}($-32.7\%$)}
    & --
    &
    & -- \\

    \midrule

    Img-pres.
    & Joint WAM $\rightarrow$ IDM WAM
    & 98.1
    & 98.4
    & 67.0 {\scriptsize\color{gray}($-31.4\%$)}
    & $\longrightarrow$
    & 71.3 {\scriptsize\color{gray}($-27.1\%$)}
    & 14.18
    & $\longrightarrow$
    & 13.85 \\

    Img-pres.
    & IDM WAM $\rightarrow$ Joint WAM
    & 98.4
    & 98.1
    & 63.0 {\scriptsize\color{gray}($-35.1\%$)}
    & $\longrightarrow$
    & 65.0 {\scriptsize\color{gray}($-33.1\%$)}
    & 14.24
    & $\longrightarrow$
    & 14.16 \\

    \hline
    \bottomrule
\end{tabular}%
}
\vspace{-4mm}
\end{table*}

\subsection{Transferability Across WAM Variants}
\label{sec:eval-transfer}

To answer \textbf{RQ6}, Table~\ref{tab:transferability-results} compares attacks transferred from a source WAM with the same objective optimized directly on the target WAM. For action-only attacks, transferred task success is 64.1\% for Action-only WAM $\rightarrow$ Joint WAM, 70.8\% for Joint WAM $\rightarrow$ IDM WAM, and 65.4\% for IDM WAM $\rightarrow$ Joint WAM. These values are only 2.6, 4.7, and 3.9 percentage points (pp) higher, respectively, than attacking the target directly, while still reducing target clean success by 27.6--34.0 (pp).

Imagination-preserving attacks also transfer effectively between the Joint and IDM WAMs. Joint WAM $\rightarrow$ IDM WAM transfer yields 71.3\% target success, compared with 67.0\% under a direct attack, while IDM WAM $\rightarrow$ Joint WAM transfer yields 65.0\%, compared with 63.0\% under a direct attack. The corresponding direct/transferred predicted-future distances are 14.18/13.85 and 14.24/14.16, respectively, showing that the perturbations retain comparable imagination-preserving behavior after transfer. Together with the action-only results, transferred attacks are only 2.0--4.7 pp weaker than direct attacks across all five directions. This consistent transferability suggests that BadWAM exploits shared observation-space vulnerabilities across WAM variants rather than overfitting to a particular action head.

\subsection{What Do These Results Imply for WAM Safety?}
\label{sec:eval-implications}

Across \textbf{RQ1-RQ6}, the results point to a subtle failure mode for imagination-based safety checks. The relevant safety property is not the predicted-future output in isolation, but synchronization between the imagined future and the action that will actually be executed.

This also explains why the attack is WAM-specific. For a conventional action-only policy, an adversarial perturbation can only be assessed through its effect on the action or final task outcome. For a WAM, the attacker can exploit a gap between two interfaces: the future-imagination interface and the action interface. The action-only BadWAM objective targets the high-strength endpoint of this problem; the imagination-preserving objective targets a stealthier endpoint. The latter is especially relevant for systems that expose predicted futures to planners, runtime monitors, or human supervisors. Even when the predicted future remains visually close to the clean prediction, the executed action chunk may no longer be the action that would realize that future.

Finally, the transfer results suggest that the vulnerability is not merely an implementation artifact. Perturbations optimized on one WAM variant remain damaging on another, indicating shared observation-space directions that affect action generation.

\section{Conclusion}

World-action models promise a tighter connection between perception, prediction, and control by coupling action generation with imagined futures. In this paper, we show that this coupling can itself become an attack surface. We introduce BadWAM, a black-box framework for world-action drift attacks that induces task failures through small observation perturbations. Beyond action-only attacks, BadWAM exposes a more WAM-specific failure mode: action-imagination decoupling, where the executed actions are shifted while its predicted future remains comparatively stable. Across multiple WAM variants, BadWAM substantially reduces task success under bounded perturbations.

\bibliographystyle{plain}
\bibliography{reference}

\newpage
\appendix

\twocolumn[
\begin{center}
    {\Large\bf Appendix}
\end{center}
\vspace{1em}
]

\section{Details of Experiment Setup}

\begin{table*}[t]
    \centering
    \caption{
    Training configurations of the four WAMs.
    ``Video input'' denotes the spatial resolution after composing all camera
    views, and $d_a/d_p$ denotes the action/proprioception dimensions.
    ``Seq./horizon'' reports the number of frames and the corresponding action
    horizon.
    The effective batch size is
    $B_{\mathrm{eff}}=B_{\mathrm{GPU}}\times 8\times\mathrm{Acc.}$
    }
    \label{tab:training-configurations}
    \resizebox{\textwidth}{!}{%
    \begin{tabular}{llccccccccccc}
        \toprule
        Dataset
        & Model
        & Views
        & Video input
        & $d_a/d_p$
        & Seq./horizon
        & Epochs
        & Opt. steps
        & $B_{\mathrm{GPU}}$
        & Acc.
        & $B_{\mathrm{eff}}$
        & Normalization
        & Grad. ckpt. \\
        \midrule

        LIBERO
        & Joint WAM
        & 2
        & $224{\times}448$
        & $7/8$
        & $33/32$
        & 10
        & 21,700
        & 16
        & 1
        & 128
        & Min0max
        & No \\

        LIBERO
        & IDM WAM
        & 2
        & $224{\times}448$
        & $7/8$
        & $33/32$
        & 10
        & 21,700
        & 8
        & 2
        & 128
        & Min0max
        & Yes \\

        RoboTwin
        & Joint WAM
        & 3
        & $384{\times}320$
        & $14/14$
        & $33/32$
        & 5
        & 234,830
        & 16
        & 1
        & 128
        & Z-score
        & No \\

        RoboTwin
        & IDM WAM
        & 3
        & $384{\times}320$
        & $14/14$
        & $33/32$
        & 5
        & 234,830
        & 16
        & 1
        & 128
        & Z-score
        & Yes \\

        \bottomrule
    \end{tabular}%
    }

    \vspace{1mm}
    \begin{minipage}{0.98\textwidth}
        \footnotesize
        All models are initialized from the same Wan2.2-TI2V-5B and
        ActionDiT backbones and trained on eight NVIDIA H100 80GB GPUs
        using DeepSpeed ZeRO-1 and BF16 precision.
        We use AdamW with $\beta_1=0.9$, $\beta_2=0.95$,
        learning rate $10^{-4}$, weight decay $10^{-2}$,
        and gradient clipping at 1.0.
        The learning rate follows a cosine schedule after a 5\% linear
        warm-up and decays to $10^{-6}$.
        We set the action-loss weight to 1.0, the action-to-video frequency
        ratio to 4, and the random seed to 42.
    \end{minipage}
    \vspace{-3mm}
\end{table*}

\subsection{Training Details}
\label{training_details}

\begin{figure}[t]
    \centering
    \begin{subfigure}{0.49\columnwidth}
        \centering
        \includegraphics[width=\linewidth]{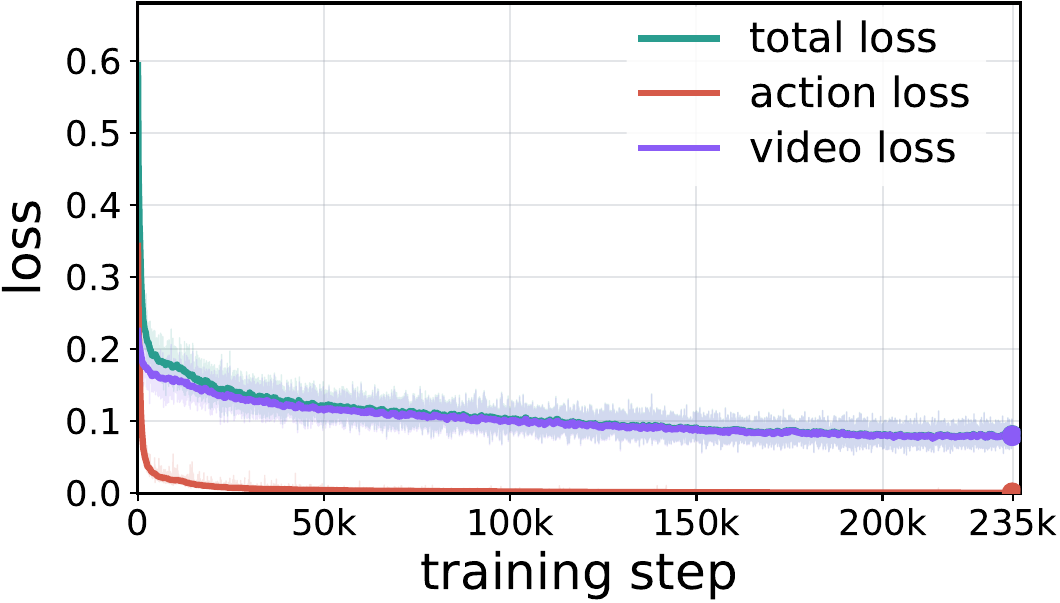}
        \caption{Linear scale.}
        \label{fig:robotwin-joint-loss-linear}
    \end{subfigure}
    \hfill
    \begin{subfigure}{0.49\columnwidth}
        \centering
        \includegraphics[width=\linewidth]{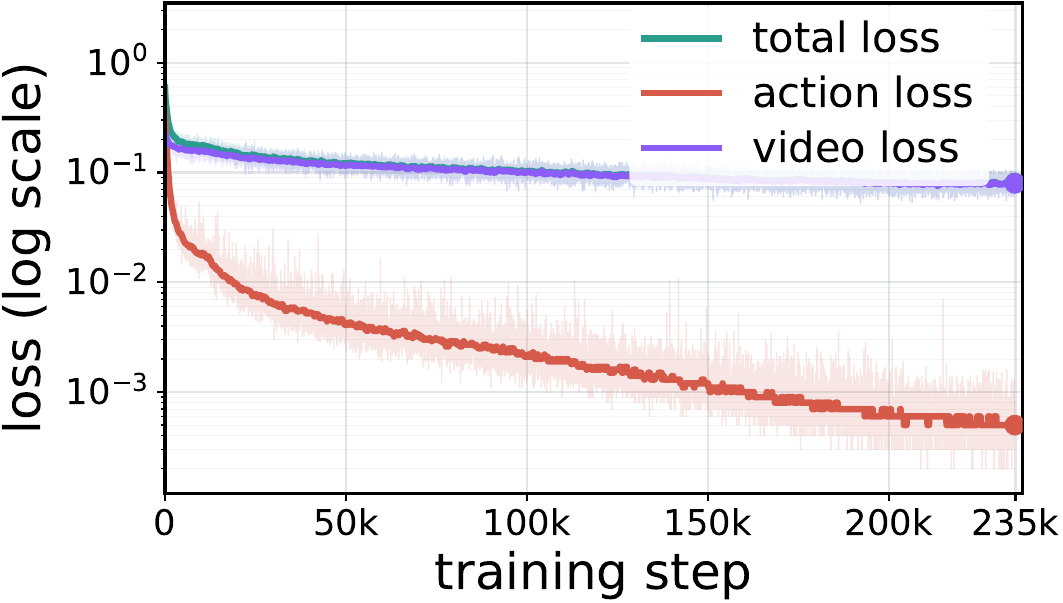}
        \caption{Log scale.}
        \label{fig:robotwin-joint-loss-log}
    \end{subfigure}
    \caption{
    RoboTwin Joint WAM training loss.
    The linear-scale view shows the overall convergence trend, while the log-scale view makes the continued decay of the action loss visible.
    Light curves denote raw logged losses and bold curves denote smoothed losses.
    }
    \label{fig:robotwin-joint-training-loss}
    \vspace{-2mm}
\end{figure}

\begin{figure}[t]
    \centering
    \begin{subfigure}{0.49\columnwidth}
        \centering
        \includegraphics[width=\linewidth]{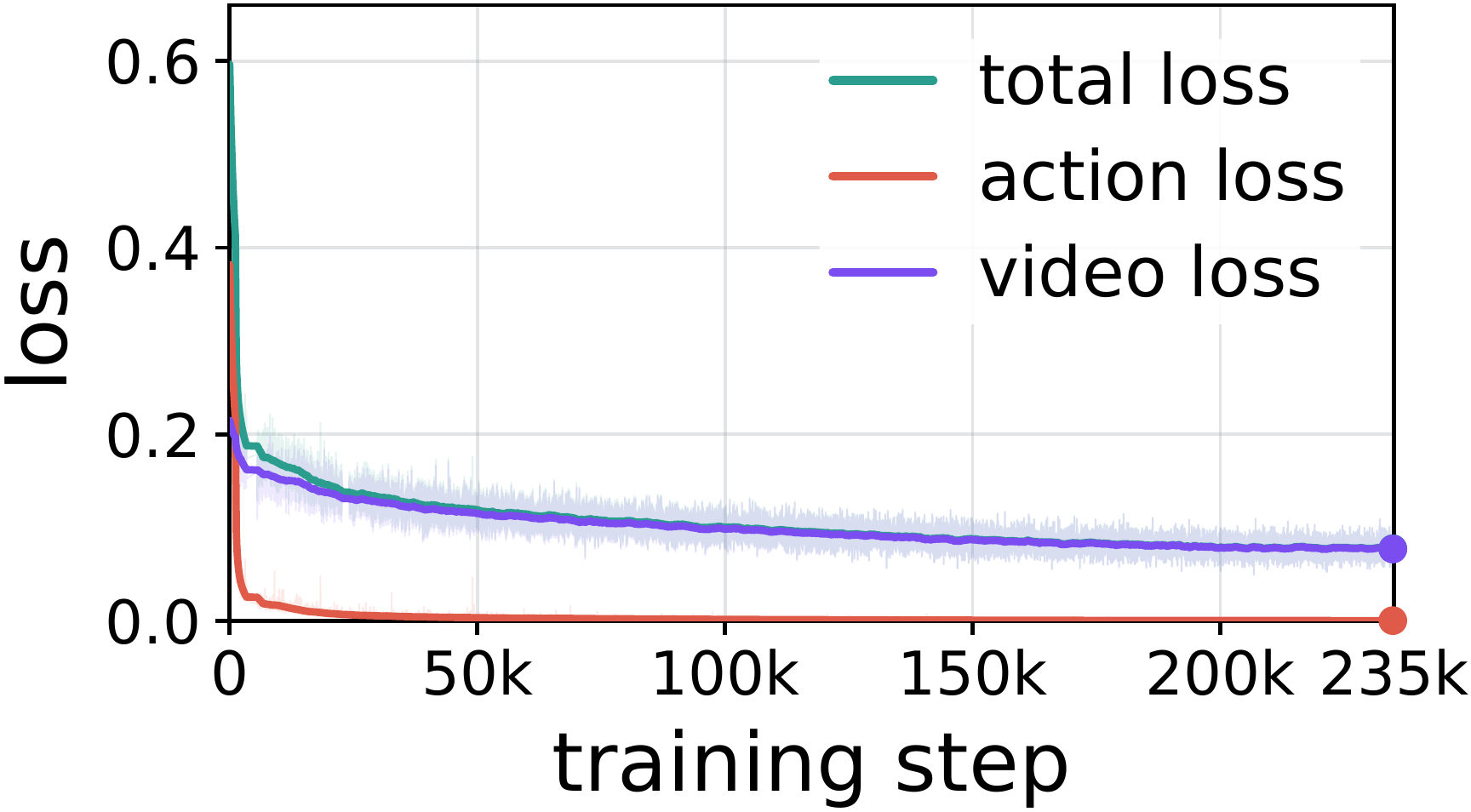}
        \caption{Linear scale.}
        \label{fig:robotwin-idm-loss-linear}
    \end{subfigure}
    \hfill
    \begin{subfigure}{0.49\columnwidth}
        \centering
        \includegraphics[width=\linewidth]{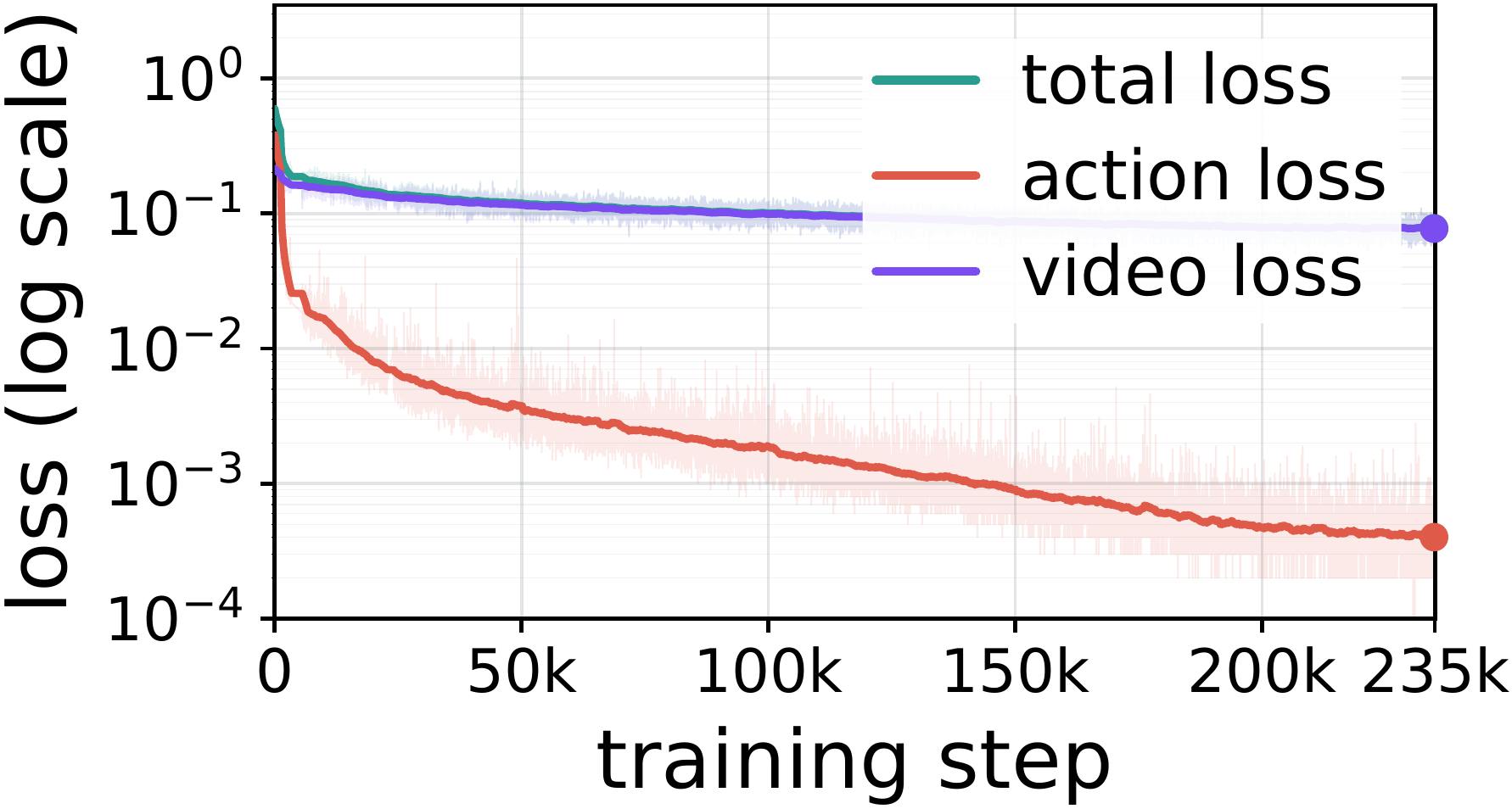}
        \caption{Log scale.}
        \label{fig:robotwin-idm-loss-log}
    \end{subfigure}
    \caption{
    RoboTwin IDM WAM training loss.
    The linear-scale view shows the overall convergence trend, while the log-scale view makes the continued decay of the action loss visible.
    Light curves denote raw logged losses and bold curves denote smoothed losses.
    }
    \label{fig:robotwin-idm-training-loss}
    \vspace{-2mm}
\end{figure}

For the action-only WAM, we use the official FastWAM checkpoints~\cite{yuan2026fastwam}. The joint
WAM and IDM WAM variants used in our LIBERO and RoboTwin experiments are trained from the FastWAM release using the default training configuration.  All the models are trained on 8 H100 GPUs. Figures~\ref{fig:robotwin-joint-training-loss} and \ref{fig:robotwin-idm-training-loss} summarize the optimization trajectories of the RoboTwin Joint and IDM WAMs, respectively.

The training setup is given in Table~\ref{tab:training-configurations}. For LIBERO, the model uses two cameras, namely the third-person view and the wrist view.  Each camera is resized to $224\times224$, and the two camera streams are concatenated horizontally. For RoboTwin, following the official FastWAM preprocessing, we compose the three camera views into a single $384\times320$ frame. The high-camera view is resized to $256\times320$ and placed above the two wrist views, each resized to $128\times160$ and concatenated horizontally.

The model backbone and tokenizer follow FastWAM.  Specifically, the configuration uses \texttt{Wan-AI/Wan2.2-TI2V-5B} as the video backbone and \texttt{Wan-AI/Wan2.1-T2V-1.3B} as the tokenizer model.  We keep the default FastWAM action and video schedulers.  We do not modify the FastWAM architecture when training the joint and IDM variants.

\subsection{Attack Hyperparameters and Query Accounting}

Unless otherwise stated, BadWAM perturbs the full preprocessed visual input under an $\ell_\infty$ bound of $\epsilon=0.06$. The perturbation is applied online at each replanning step during closed-loop execution. The default zeroth-order finite-difference search uses 16 perturbation queries plus one clean reference query per attacked replan, for a total of 17 WAM forward queries per attacked replan. Internally, the 16 perturbation queries are organized as 8 paired finite-difference updates. Each update samples one random Rademacher direction, evaluates positively and negatively perturbed inputs, and updates the perturbation by a projected signed step. We use a finite-difference radius of 0.02 and a search step size of 0.02.

For the action-only adversarial attack, the future-preserving coefficient is set to zero and the objective only maximizes clean0attacked action deviation. For the imagination-preserving adversarial attack, the objective additionally penalizes predicted-future drift.  The default full-sweep setting uses $\lambda=0.015$ for the future-preserving coefficient. The matched-strength analysis in Figure~\ref{fig:matched-stealth-dynamics} uses a larger future-preserving coefficient when comparing the two objectives under similar attack strength, so that the effect of the preservation term is easier to isolate.

\subsection{White-Box Gradient-Access Reference}

Figure~\ref{fig:random-baseline} includes a white-box gradient-access reference to contextualize the strength of our query-based attacks. This reference is not part of the main BadWAM threat model. BadWAM assumes query access and does not require gradients or model parameters. The white-box reference instead asks how much stronger the attack can become if the adversary can backpropagate through the WAM's action output.

At the beginning of an episode, the attacker optimizes one bounded input perturbation using projected gradient ascent. The objective maximizes the deviation between the clean action chunk and the perturbed action chunk. The perturbation is initialized randomly inside the $\ell_\infty$ ball, updated for 16 gradient steps with step size 0.01, and projected back to $\|\delta\|_\infty\le 0.06$ after each step. After the perturbation is found, the policy is evaluated with the standard closed-loop inference procedure. No model weights are modified, and the environment dynamics are not differentiated through.

This reference serves two purposes. First, it separates generic visual fragility from targeted action manipulation: random perturbations under the same budget are substantially weaker than targeted attacks.  Second, it shows that gradient access can further amplify the vulnerability, which suggests that the query-based BadWAM attacks are not saturating the worst-case attack strength. Since this reference uses stronger access than BadWAM, we treat it as a diagnostic upper-bound reference rather than as the primary threat model.

\subsection{Random Perturbation Baseline}

The random perturbation baseline uses the same $\ell_\infty$ budget as BadWAM but removes the optimization objective.  For each attacked observation, it samples an independent uniform perturbation from $[-\epsilon,\epsilon]$ and clips the perturbed input to the valid model input range.  This baseline is model-agnostic and does not use action outputs, future predictions, gradients, or task feedback.  Its role is to test whether task failures are caused merely by arbitrary visual corruption.  The results in Figure~\ref{fig:random-baseline} show that optimized world-action drift attacks are consistently stronger than random perturbations under the same visual budget.

\subsection{Evaluation Metrics}

Our primary metric is closed-loop task success. Lower success under attack indicates a stronger attack. In addition to success rate, we report action distance, predicted-future distance, and input perturbation statistics. The action distance is computed as the mean $\ell_2$ deviation between clean and attacked action chunks. The predicted-future distance is computed analogously on the future representation or decoded future-video tensor available through the WAM interface. The decoupling score is the ratio between action distance and predicted-future distance, with a small numerical constant in the denominator for stability. A high decoupling score indicates that the attack obtains a larger action shift per unit of predicted-future drift.

For repeated-trial evaluation, we use pass@$k$. For a fixed task, pass@$k$ is the fraction of successful executions among the first $k$ trials. We then average this quantity across tasks. This metric shows whether an attack only causes isolated unlucky failures or consistently lowers reliability as more trials are considered.

\end{document}